\documentclass[10pt,twocolumn,letterpaper]{article}

\usepackage[pagenumbers]{cvpr} 

\usepackage[dvipsnames]{xcolor}

\definecolor{cvprblue}{rgb}{0.21,0.49,0.74}
\usepackage[pagebackref,breaklinks,colorlinks,citecolor=cvprblue]{hyperref}
\usepackage{multirow}

\usepackage{xcolor}

\def\paperID{15713} 
\def\confName{CVPR}
\def\confYear{2024}

\title{Visual Objectification in Films: Towards a New AI Task for Video Interpretation}

\author{Julie Tores$^{1,2}$ \hspace{.5cm} Lucile Sassatelli$^{1,3}$\hspace{.5cm}   Hui-Yin Wu$^{4}$ \hspace{.5cm} Clement Bergman$^{4}$ \\ Léa Andolfi$^{7}$  \hspace{.5cm}
Victor Ecrement$^{4}$ \hspace{.5cm} Frédéric Precioso$^{2}$ \hspace{.5cm} Thierry Devars$^{6}$ \\ Magali Guaresi$^{5}$ \hspace{.5cm} Virginie Julliard$^{6}$ \hspace{.5cm} Sarah Lecossais$^{7}$\\
  $^{1}$Université Côte d’Azur, CNRS, I3S, France \hspace{.5cm}
  $^{2}$Université Côte d’Azur, CNRS, Inria, I3S, France \\
  $^{3}$Institut Universitaire de France \hspace{.5cm}
  $^{4}$Université Côte d'Azur, Inria, France\\
  $^{5}$Université Côte d'Azur, CNRS, BCL, France \hspace{.5cm}
  $^{6}$Sorbonne Université, GRIPIC\\
  $^{7}$Université Sorbonne Paris Nord, LabSIC\\
  {\tt\small {julie.tores@univ-cotedazur.fr}}
 }

\begin{document}
\maketitle

\begin{abstract}

In film gender studies, the concept of “male gaze” refers to the way the characters are portrayed on-screen as objects of desire rather than subjects. In this article, we introduce a novel video-interpretation task, to detect character objectification in films. The purpose is to reveal and quantify the usage of complex temporal patterns operated in cinema to produce the cognitive perception of objectification.
We introduce the ObyGaze12 dataset, made of 1914 movie clips densely annotated by experts for objectification concepts identified in film studies and psychology.
We evaluate recent vision models, show the feasibility of the task and where the challenges remain with concept bottleneck models. 
Our new dataset and code are made available to the community.
\end{abstract}

\section{Introduction}\label{sec:intro}

\begin{figure*}[t]
  \centering
  \includegraphics[width=0.9\linewidth]{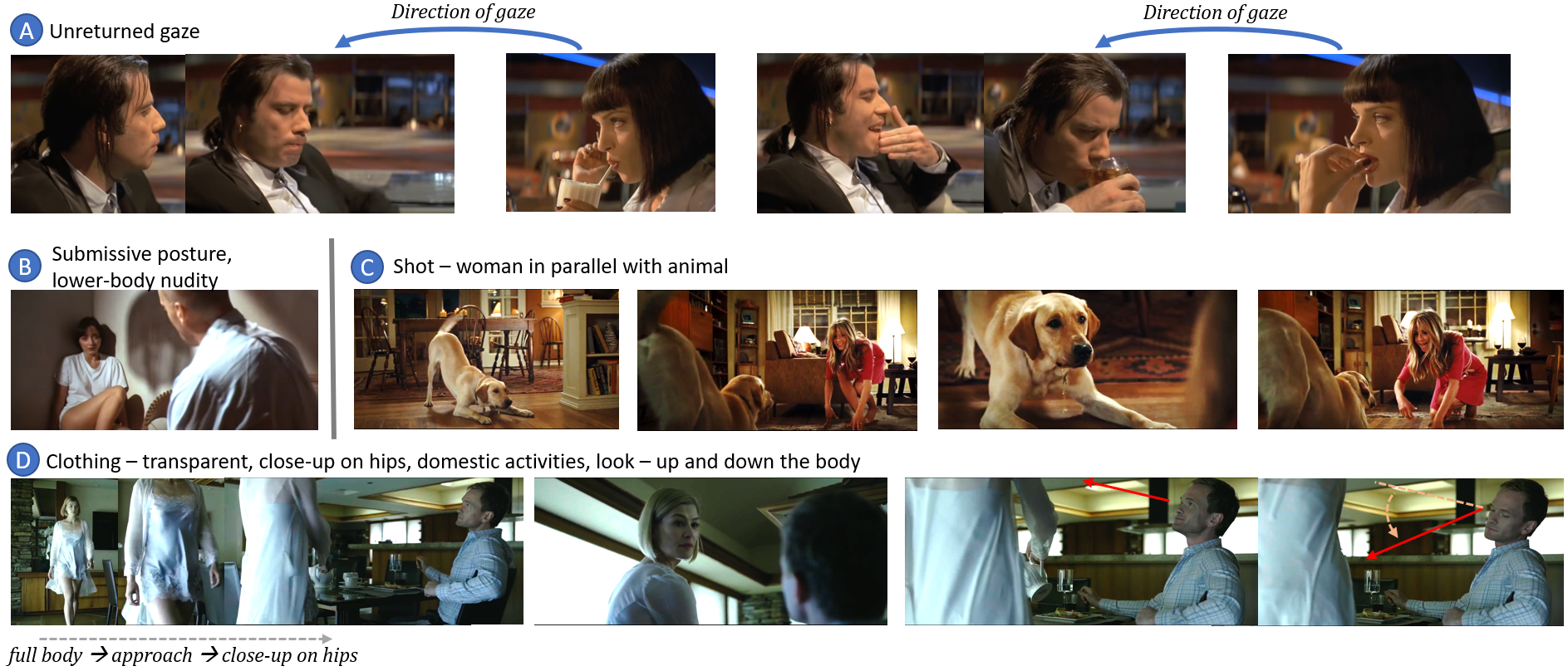}
  \caption{In modern film media, the unequal characterization of gender on screen frequently evokes concepts of objectification, such as (A) unequal gaze (\textit{Pulp Fiction}, 1994), (B) Nudity and submissive postures (\textit{Pulp Fiction}, 1994), (C) animalisation or infantilisation (\textit{Marley and Me}, 2008), and (D) transparent clothing, camera framing, domestic gender roles, and voyeurism (\textit{Gone Girl}, 2014).}
  \label{fig_gaze}
\end{figure*}

In film gender studies, the concept of “male gaze” \cite{mulvey1975} refers to the way the characters – especially women – are portrayed on-screen as objects of desire rather than subjects. Consider in Figure~\ref{fig_gaze} how objectification is manifested in various ways such as how the camera is placed and moved, the gaze interactions between characters, the choice of clothing, and arrangement of scene elements.
Such disparities in how people are presented, depicted or addressed to in digital contents based on their gender has large-scale social implications such as the perpetuation of harmful stereotypes and hostile social situations. 

These disparities have been the subject of an increasing number of studies at the intersection of social and computational sciences. In online social networks, computational approaches to sexism detection have been increasingly investigated for textual data, as a part of hate speech detection. As explained by Samory et al. \cite{samory_call_nodate}, sexism is a complex sociological construct, whose high-level interpretive nature and subtle dimensions beyond offensive speech make for an unsolved challenge.

In visual media such as films and TV series, the characters they depict shape our collective imagination and perception of sociological constructs, such as gender, race, and class. Currently, most large-scale approaches to understanding gender representation in these media have focused on quantifying the presence of women in the image and audio content (e.g., \cite{GMMP,somandepalli_computational_2021,mazieres_computational_2021}). However, works in social sciences show that quantifying the presence of gender on screen is insufficient for grasping the issue of gender inequalities in visual media.
For films, the classic Bechdel test, although useful and simple, considers neither the visual modality, which is key to analyzing gender depiction \cite{mulvey1975}, nor the textual constructs of speech and dialogue \cite{schofield_gender-distinguishing_2016}. 
While a few works have investigated sexist memes \cite{fersini_detecting_2019,gasparini_benchmark_2021}, sexist advertisement \cite{gasparini_multimodal_2018}, and characterized the on-screen positioning and co-occurrence of certain groups with respect to scene types and objects (e.g., Wang et al. for still images \cite{vedaldi_revise_2020}, Jang et al. \cite{jang_quantification_2019} for films), computational approaches to interpretive sexism in visual media remain very scarce.

In this article, we introduce a new challenging task for computer vision: detecting character objectification in films.
We present a major step to tackle the question of subtle sexism in videos, operationalizing the popularly known concept of male gaze with the construct of objectification, and specifically considering the temporal dimension where such video patterns unfold.
The end purpose is to enable large-scale quantification and characterization of complex patterns producing on-screen objectification, and unveil possible correlations along the lines of the gender or race constructs. Owing to the importance of this question, we consider it is critical to support the design of explainable methods and fine-grained model error analysis, which we address by densely annotating video data for theory-driven concepts.\\

\textbf{Our contributions are}:\\
\noindent $\bullet$ We introduce a novel video-interpretation task, to detect character objectification in films. This is an interpretive task, hence extending beyond the more classical yet still challenging video-understanding tasks, and involving a subjective judgement. In a team involving media studies experts and building on results in cinematography and psychology, we design a thesaurus of visual objectification, defining coarse-grained concepts with exemplified instances. This thesaurus is then used to formulate precise annotation guidelines. We introduce the \textbf{\textit{ObyGaze12}} dataset to the community, with 1914 clips of 12 films densely annotated by experts for concepts of objectifying gaze, including hard negative examples. It corresponds to 25\% of the MovieGraphs dataset. We verify the consistency of the obtained data, and provide first analyses showing the compositional nature of objectification. The dataset is meant to explore the complex temporal patterns producing character objectification in films. To the best of our knowledge, it is the first work proposing a computational approach to this interpretive task in videos. \\
\noindent $\bullet$ We verify that the new task of objectification detection in videos is accessible by testing recent vision and vision-language models, and that hard negative examples improve classification. We also investigate the model weaknesses in representing every objectification concept. To do so a thorough analysis is carried out with Concept Bottleneck Models (CBMs) and allows us identify that the challenges specifically lie into representing the concepts of Type of shot, Look, Posture and Appearance.\\

To the best of our knowledge, this is one of the few video datasets with dense concept-based annotations for a high-level construct, and the first for objectification.
The dataset and code used in this article is entirely provided in \url{https://anonymous.4open.science/r/ObyGaze12/}.

The article is organized as follows. In Sec. \ref{sec:rel_works}, we first provide a review of the relevant works on visual biases in films, dataset creation and models for video understanding. We then introduce dataset creation for the new \textit{ObyGaze12} and present first analyses in Sec. \ref{sec:data}. Sec.~\ref{sec:exp} presents the evaluation of models on the new task, and the analysis of the difficulty of concept representation with CBMs. Finally, we provide discussions on ethical aspects and challenges in Sec. \ref{sec:discu}, and conclusions in Sec. \ref{sec:conclu}.

\section{Related works}\label{sec:rel_works}

In this section, we position our contributions with respect to the relevant existing work related to each of its key ingredients. First, we introduce biases in visual datasets and computational approaches to analysis of visual gender representation in films. We then discuss interpretive-level tasks, increasingly common in natural language processing and the approaches to dataset creation to instantiate them for ML approaches. We highlight here the scarcity of visual datasets made for high-level interpretive tasks, in particular for video data. Finally, we introduce the video understanding approaches that we consider to benchmark on our new interpretive task, specifically focusing on explainable concept-based approaches to locate the challenges ahead in video interpretation for this new task.

\subsection{Visual biases in film datasets}
The task we introduce is connected to the general problem of bias detection. As exposed by Fabbrizzi et al. \cite{fabbrizzi_survey_2022}, biases in visual datasets can be classified into selection bias (how subjects are included in a dataset), framing bias (how the visual content has been artificially composed) and label bias (errors or disparities in the labelling data). Our contribution is closely related to the framing and labelling biases.

In film datasets, the first studies of biases in gender representation were from a presence quantification perspective. Guha et al. \cite{guha_gender_2015} and Somandepalli et al. \cite{somandepalli_computational_2021} automatically estimate the screen time (from video) and speaking time (from audio) of male and female characters in Hollywood movies. They show that women are seen (36\% screen time) and heard (41\% screen time) significantly less than male characters. Analyzing audiovisual co-occurrence, they also show that male faces are more likely to appear on screen even when a female character is speaking, while male and female speech have equal probability when a female character appears. Mazieres et al. \cite{mazieres_computational_2021} carry out such a quantitative analysis with a movie dataset spanning three decades, and show a temporal trend towards general fairer representation between both binary genders. They however show that the applied framings remain unfair, with only 40\% of one-face frames featuring a female (60\% for males). Jang et al. specifically analyze the qualitative framing differences in gender portrayal in a dataset of 20 Hollywood movies and 20 Korean movies \cite{jang_quantification_2019}. They show that female characters are portrayed with lower emotional diversity, spatial occupancy, temporal occupancy, intellectual image or mean age. From character-object co-occurrence, they also showed that female characters appear indoors and are described in a less dynamic way than male characters.

In contrast in this article, we take a first step towards detecting bias in gender representation from a high-level construct, objectification, qualitatively described in various disciplines such as cinematography \cite{mulvey1975,brey2020regard}, social psychology \cite{calogero_operationalizing_2011,kozee_interpersonal_2011} and neuroscience \cite{bernard_revealing_2019,bernard_objectifying_2018,bernard_sexualizationobjectification_2020}.
Objectification is produced by complex temporal patterns never analyzed computationally in videos until now.

\subsection{Interpretive-level tasks and dataset creation}
At the same time that biases in visual datasets are uncovered and analyzed, other approaches aim to detect bias in human data. 
Detecting highly interpretive constructs, such as hate speech, propaganda, sexism, racism, has been a long-standing endeavor in NLP. These constructs involve multiple dimensions often spanning several disciplines.
Da San Martino et al. \cite{da_san_martino_fine-grained_2019} considered the three-fold difficulty of propaganda detection: (1) deciding generic propaganda techniques that can be used to produce dense annotations on standalone news articles, (2) obtaining annotation not at the level of the entire article but at a finer-grained level with tagging of annotator-decided text spans, and (3) working around crowdsourced annotators who are heavily influenced by their political views. They resort to 4 experts to annotate new articles with text spans to be associated to 18 possible propaganda techniques. Samory et al. uncover a similar challenge of construct complexity for sexism detection \cite{samory_call_nodate}. They observe that multiple articles for automating this task consider widely different definitions of sexism, often referring to sub-dimensions of the broader construct.  
This makes it difficult to properly compare and assess the models, specifically identifying the actual dimensions they capture. Samory et al. consider existing works in social psychology where sexism dimensions have long been operationalized with sub-scale questions tested for consistency.
To approach dataset creation for on-screen objectification, we inspire from these two last works, and on dense annotation approaches of image datasets recently proposed for the medical domain \cite{daneshjou2022skincon}.
To the best of our knowledge, approaches for sexism detection from visual content are scarce, and almost none existent for video data. Two main types of visual content have been considered so far: hateful or sexist memes \cite{kiela_hatefulmeme_2020,fersini_detecting_2019}, and sexist advertisements \cite{gasparini_multimodal_2018} which can also relate to symbolic advertisement understanding \cite{hussain_automatic_2017, kalanat_symbolic_2022}.

In films, analysis of biases in gender representation has been also automated with NLP approaches applied to film scripts. Agarwal et al. \cite{agarwal_key_2015} present an approach to automatize the reference Bechdel test, made of three specific questions to assess whether women are portrayed as less important characters in a film. Martinez et al. \cite{martinez_boys_2022} propose a RNN-based model to automatically extract agent-verb-patient triplets. From 912 movie scripts, they show that male characters are associated with a higher agency while female characters are more frequently the object of gaze.
Apart from these two works analyzing character and action co-occurrences, which already go beyond mere presence or speaking time, Su et al. introduce the more abstract task of trope understanding in movies \cite{su_truman_2021}. Tropes are storytelling devices conveying abstract concepts whose instantiation through time and space can vary widely. Their analysis hence goes beyond shallow video event understanding and instead requires deep cognition skills. Our work is close to this last one as we also introduce a dataset of films annotated for a higher-level construct beyond event and story understanding. However, we provide dense annotations of sequences with constitutive concepts for the high-level construct of character objectification, whose instantiation can also widely vary with the filmmaker. Also, while Su et al. exploit an existing online base contributed to by the community, we design and carry out a strictly defined annotation process by experts.

In this article, we set out from the concept of male gaze defined in various ways in film gender studies. Mulvey \cite{mulvey1975} characterizes the concept of gaze by the three relations between the camera and the characters, between the characters, but also between the spectator and the characters, involving a possible film-external cognitive component. While some formalizations of gaze set out from the gender of the director \cite{Mal18}, Brey defines female and male gaze only from the film content, with a corpus analysis focusing on aesthetics \cite{brey2020regard} and revolving around the construct of objectification. 

We start here from the latter cinematographic analysis of male gaze, build on its operationalization in both cinematography and psychology, to create the conditions to make a new challenge accessible to the CV community: we produce a strict annotation process to obtain densely annotated video data and analyze where the new challenges lie ahead. We generally position our approach producing a non-large scale but high-quality dataset we hope useful within the lines of the call of  Paullada et al. for such data-centric AI approaches \cite{paullada_data_2021}.

\subsection{Approaches to video and movie understanding}

\paragraph{Pre-trained models for video understanding}
Cross-modal foundation models, such as CLIP \cite{CLIP} and ALIGN \cite{ALIGN}, learn aligned image and text representations through contrastive pre-training on large-scale closed datasets. These vision-language transformers have brought major improvements to few-shot and zero-shot cross-modal and vision tasks, such as Visual Question Answering (VQA) or image classification and segmentation \cite{clip_is_also_an_efficient}. 

Generalizations of the CLIP model to video data have included VideoCLIP \cite{xu_videoclip_2021} and two X-CLIP models \cite{X-CLIP,XCLIP}. 

In particular, X-CLIP \cite{XCLIP}, which we employ in this article, expands CLIP with video temporal modeling and video-adaptive textual prompts.
Adaptation of foundation models to end tasks and domains is key to their success and an active area of research. Prompt tuning in particular is an approach to learn input data perturbations so that a frozen model can perform a new task. Visual prompt tuning has been recently introduced \cite{zhou_learning_2022,bahng_exploring_2022}, investigated \cite{PromptCAL,maple, zhou_conditional_2022} and explored for various tasks such as text-video retrieval \cite{VoP} and multimodal tracking \cite{visual_prompt_multimodal_tracking}.

\paragraph{Movie-related tasks}
Legacy and large-scale pre-trained vision-language models have been leveraged for movie-related tasks. Bose et al. consider the difficulty of visual scene recognition in movies due to the domain mismatch between training frames and scene images, and create the MovieCLIP dataset obtained from weakly labeling movie shots from scene categories using the CLIP model \cite{MovieCLIP}.
An important vision-language movie-related task is audio-description for the visually-impaired, for which a major dataset introduced recently is MAD \cite{MAD}, gathering sparse natural language sentences grounded in over 1200 hours of movie videos. AutoAD \cite{han_autoad_2023} and AutoADII \cite{Han_2023_ICCV} are two recent approaches to generate audio-description from the video, both leveraging CLIP to learn to prompt GPT. 

To design approaches to learn human-level constructs, such as emotions, interactions or relationships, datasets generated with human supervision are also instrumental. A prominent representative is the MovieGraphs dataset, providing detailed annotations of clips of 51 movies with emotional states, character interactions and relationships, and other scene reasoning elements \cite{vicol_moviegraphs_2018}. It has prompted works tackling such recognition tasks \cite{Gan2023CollaborativeNL, kukleva_learning_2020}. In this article, we build on the MovieGraphs dataset to annotate a selection with the construct of objectification, to later analyze it in connection with the other annotated social elements.

\paragraph{Concept-based models}
In this article, we aim to assess the capacity of CLIP-based methods to provide relevant embeddings for the concept of objectification. We do so with a direct evaluation of classification results when an adapter (MLP) fed by X-CLIP embeddings is learnt. We analyze the results with a concept-based approach by building on Concept Bottleneck Models (CBM) \cite{pmlr-v119-koh20a}. 
Concept-based models are an active area of research in XAI, with works tackling the accuracy-explainability tradeoff \cite{zarlenga2022concept} and the need for user-defined concepts \cite{yang_language_2023_labo}.
We specifically employ Post-hoc CBM (PCBM) \cite{yuksekgonul2022posthoc}, which consists in learning a concept subspace (made of Concept Activation Vectors \cite{Kim2017InterpretabilityBF}) in the embedding space of the pre-trained model. Data samples are then projected in this concept subspace, from where the classification task can be performed with an interpretable classifier.

\section{Data and methods}\label{sec:data}

This section presents our approach to create the first dataset for visual objectification in videos, specifically in films. We name this dataset \textit{ObyGaze12}, short for \textit{ObjectifyingGaze12}, which has the following \textbf{highlights}:\\
\noindent$\bullet$ It considers the multiple dimensions of the construct of visual objectification, made of filmic (framing and editing over successive shots, camera motion, etc.) and iconographic properties (visible objects, body parts, attire, character interactions, etc.).\\
\noindent$\bullet$ It is based on a thesaurus articulating five sub-constructs identified from multidisciplinary literature (film studies, psychology) from which we define typical instances, then grouped into coarse-grained visual concepts.\\
\noindent$\bullet$ The data is annotated densely with concepts, and shows the multi-factorial property of objectification, corroborating with some recent developments in cognitive psychology \cite{bernard_revealing_2019}.\\
\noindent$\bullet$ Categories to annotate include a hard negative category meant to perform fine-grained error analysis and improve model generalization.

\subsection{A thesaurus of objectification}

We first formalize the construct of visual objectification and derive key concepts to annotate in film scenes.

Together with media studies experts, we identify five sub-constructs of objectification from literature on film cognition and film gender studies, and social and cognitive psychology: male gaze (point of view of a man on a woman) \cite{brey2020regard,mulvey1975,bernard_objectifying_2018}, sexualization \cite{bernard_sexualizationobjectification_2020,bernard_revealing_2019}, surveillance of the feminine body \cite{calogero_test_2004,mckinley1996objectified,denchik2005development}, female inaction / male possession \cite{gervais_social_2020,sap2017connotation}, and infantilism / animalization \cite{mulvey1975}.

These sub-constructs come with typical instances and examples from filmmaking techniques (\cite{mulvey1975,brey2020regard}) or validated questionnaires (\cite{calogero_test_2004, mckinley1996objectified}), as shown in the middle and right-most columns in Table~\ref{tab-objthesaurus}. These typical instances are then grouped into eight coarse-grained visual concepts, corresponding to the possible means of production of visual objectification.
They are shown in the left-most column in  Table~\ref{tab-objthesaurus}: type of shot (framing and gaze of camera), look (gaze of characters on the other), body (partial or full nudity, and sexually suggestive body parts), posture (connoting, e.g., childhood, submission or inaction), clothing (in relation to context and activities), appearance (age and makeup), expression of emotion (restrained or exaggerated according to gender role), and activities (linked to gender roles).
Visual examples are provided in Fig.~\ref{fig_gaze}, where we show video samples of objectification concepts \textit{Look}, \textit{Posture}, \textit{Type of shot}, \textit{Clothes} and \textit{Activity}.

\begin{table*}
\footnotesize
\caption{Thesaurus of the typical instances and examples of visual objectification in films, grouped into eight main visual concepts used for annotation. Examples are possible means to produce one of the five sub-constructs of objectification (male gaze, sexualisation, surveillance of the feminine body, female inaction/male possession, infantilism/animalisation), to be assessed by annotators.}
\label{tab-objthesaurus}
\begin{tabular}{|p{2.1cm}|p{5.5cm}|p{8.5cm}|}
\hline
Concept                        & Concept instances   & Examples                                                                                       \\ \hline
\multirow{3}{*}{Type of shot} & Shot suggesting man perspective in presence of woman                                    & close-up on a man's face; body parts of woman                   \\ \cline{2-3}                                    
                              & Shot suggesting man gaze on woman                                    & camera takes the perspective of a male character with first close-up on the face followed by camera motion looking a woman from bottom up                                        \\ \cline{2-3} 
                              & Shot showing a woman in parallel with an animal                   & woman at same level and position with a dog            \\ \hline

\multirow{3}{*}{Look}         

                              & Voyerism                                                                   & character watching another one without their knowledge                                               \\ \cline{2-3}                                         
                              & Non-reciprocal gaze                                                      & woman looking at man who does not look back                                                    \\ \hline  

\multirow{4}{*}{Body}         & Suggested nudity                                                            & clothing on floor; silhouette behind shower curtain; nude shadow on wall             \\ \cline{2-3} 
                              & Partial nudity                                                              & nude upper or lower body; partially open clothing or draping; in underwear                     \\ \cline{2-3} 
                              & Full nudity                                                                 & nude person fully or partially shown                                                           \\ \cline{2-3} 
                              & Body parts suggestive of  sex                                               & close-up shots on breast, buttocks, hips, or lips                                              \\ \hline
\multirow{6}{*}{Posture}      & Gesture or posture connoting seduction                                      & lip-biting; hip roll; twisting or tucking hair                                 \\ \cline{2-3} 
                              & Gesture or posture connoting sexuality                                      & eating phallic symbols; arching back                                 \\ \cline{2-3} 
                              & Gesture or posture connoting inaction                                       & being undressed by someone                                                                       \\ \cline{2-3} 
                              & Gesture or posture connoting submission                                     & leaning on a man                                                                               \\ \cline{2-3} 
                              & Gesture or posture connoting dependence                                     & following a man                                                                                \\ \cline{2-3} 
                              & Skipping                                                                    & skipping gait                                                                                  \\ \hline
\multirow{4}{*}{Clothing}     & Wet or transparent clothing                                                 & thin shirt soaked in rain                                                                      \\ \cline{2-3} 
                              & Clothing impractical to situation                                           & wear pumps for running, a skirt when gardening                                           \\ \cline{2-3} 
                              & Color code associated to character                                          & woman with pink clothing and accessories                                                       \\ \cline{2-3} 
                              & Older woman wearing infantile clothing                                      & woman wearing an Alice band or high socks                                                      \\ \hline
Appearance                    & Discrepancy between appearance of woman and context or biographical elements & perfect makeup when waking up; mother of heroine too young; young girl played by older actress \\ \hline
Exp. of emotion         & Asymmetric expression of emotion                                            & boys don't cry; woman being hysterical                                                         \\ \hline
Activity                    & Doing domestic activities                                                   & doing laundry, cooking, cleaning, being constantly in the kitchen                              \\ \hline

\end{tabular}
\end{table*}

\subsection{Data selection}

Over the various existing movie datasets (see Sec. \ref{sec:rel_works} and \cite{MAD, rohrbach_movie_2017, vedaldi_movienet_2020, bain_condensed_2020, avidan_moviecuts_2022, mazieres_computational_2021,vicol_moviegraphs_2018}, \cite[Table 1]{somandepalli_computational_2021}), many have overlapping titles and only a few have rich human supervision. Amongst these, the MovieGraphs dataset \cite{vicol_moviegraphs_2018} includes rich, high-level human annotations of 7637 clips of 51 movies, with emotional states, interactions and relationships, and other social reasoning elements. 
These elements are important and valuable in exploring social concepts of objectification in visual media. 
These movies also frequently appear in \cite{vedaldi_movienet_2020, bain_condensed_2020, mazieres_computational_2021}.
The movie clips have also a short duration -- mostly within 1-5 minutes -- facilitating dense and granular annotations (over 100 clips for an average 2 hour film) while preserving the possibility to observe longer-term interactions and story development across a number of shots.
From the 7637 clips of the MovieGraphs dataset, we select 1914 clips to annotate for objectification, amounting to 25\% of the dataset and 12 complete movies, which were selected to approximately reproduce the fraction of genres in the original dataset. The list of selected films, year of release, and genre can be found in Table~\ref{tab-films} in Appendix~\ref{sec_suppl:dataset}.

\subsection{Data annotation}

Every selected movie is annotated by at least two experts for objectification level and concepts over the movie scenes. Specifically, the annotators were asked to repeat a three step process for every scene they deemed interesting from an objectification perspective: (1) watch the movie entirely and when they identify a scene worth annotating, (2) delimit the clip by using the cutting function in our annotation tool, and (3) assigning an objectification level and annotate the concept(s) involved in the objectification rating. We define four levels of objectification:

\begin{itemize}
    \item \textbf{Easy Negative}: there are no elements suggestive of objectification. No concept can be annotated. Default value for watched but non-selected scenes.
    \item \textbf{Hard Negative}: the scene contains elements of objectification from the thesaurus, but their presence does not result in an objectification effect.
    \item \textbf{Not Sure}: the scene indicates objectification, but does not completely fit the definition in the thesaurus.
    \item \textbf{Sure}: the scene contains elements of objectification from the thesaurus, and their presence results in an objectification effect.
\end{itemize}

Four expert annotators were recruited to annotate the dataset. A first presentation session was held to introduce our annotation tool and the annotation procedure. All four annotators were then given the same two films -- \textit{Juno} and \textit{Silver Linings Playbook} -- to annotate using the proposed methodology. A second meeting was then set up after the annotation of the two films to analyze the reasons for divergence and remedy them by identify which elements of the annotation guidelines to clarify and how. We then randomly assigned two annotators to each of the remaining 10 films to annotate separately. Our resulting dataset \textit{ObyGaze12} is available at \url{https://anonymous.4open.science/r/ObyGaze12/}.

\paragraph{Data processing and fusion}
The data processing is described in detail in Appendix~\ref{sec_suppl:dataset}. Following the annotation step, the annotations are then projected from the delimitations provided by each annotator onto the delimitations of the MovieGraphs clips. Since multiple annotations may overlap the same MovieGraphs clip, the annotation that is projected on the MovieGraphs clip corresponds to the annotation (including objectification level and concepts) with (1) the highest level of objectification that has at least 20\% overlap with the MovieGraphs clip, and (2) when multiple annotations exist at the same level of objectification, the annotated concepts for these annotations are aggregated. 
The same process is used to aggregate the annotations of the annotators of a same clip: the maximum objectification level is kept, with possible aggregation of concepts in case both annotators chose the same level but annotated different concepts. The merged data is shared and used in the remaining of this article.

\subsection{Analysis of the \textit{ObyGaze12} dataset}
We here comment on some interesting statistics of the resulting annotations and concepts of the 1914 clips originally delimited in the MovieGraphs dataset.

First, we verify data consistency by computing the inter-annotator agreement (IAA). Given the task of annotating timespans, we choose the $\gamma$ agreement measure introduced in \cite{mathet2015unified} (and used for, e.g., annotating text spans \cite{da_san_martino_fine-grained_2019}) owing to its consideration of temporal alignment, multiple annotators, and label classification at the same time. It attributes a score between 1 (complete agreement) and $-\infty$. A value of $\gamma \leq 0$ indicates no agreement. The computation details of the $\gamma$ metric is provided in Appendix \ref{sec_suppl:dataset}.
Considering all four categories EN, HN, NS, S, we obtain and average $\gamma=0.42$. Not considering the clips annotated Not Sure (NS), which is the uncertain and ``noisy'' class in human annotations, the IAA increases to $\gamma=0.69$. This shows the consistency of the obtained annotations despite the interpretive nature of the task.
Let us also mention that recent works improve learning approaches by considering explicitly the IAA in case of low number of annotators with moderate agreement \cite{Wei2021LearningWN, wei_aggregate_2023, bucarelli_leveraging_nodate}.

\begin{figure}[b]
  \centering
  \includegraphics[width=1.0\columnwidth]{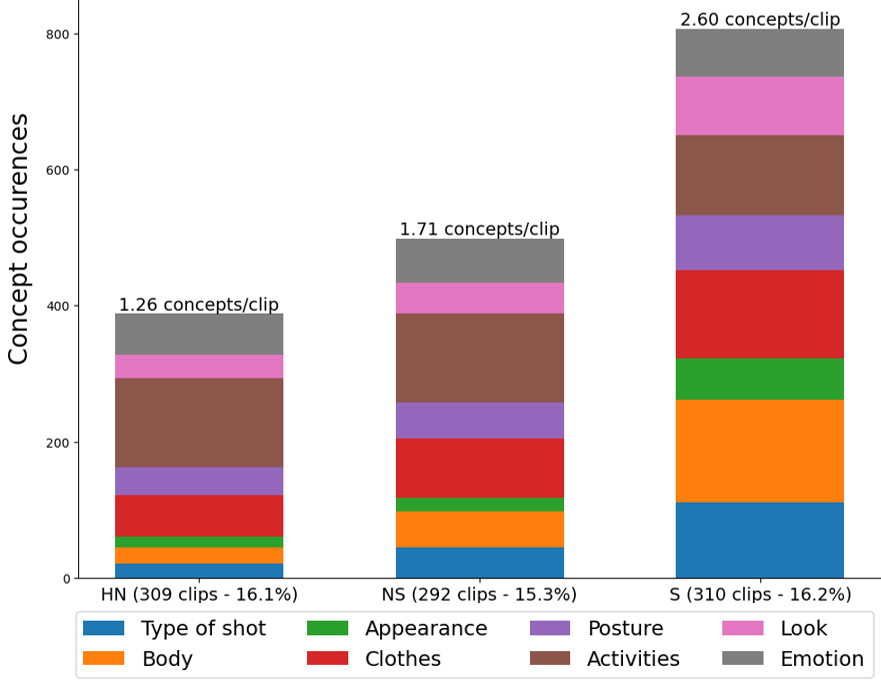}
  \caption{Distribution of visual factors annotated for each level of objectification (HN = Hard negative, NS = Not sure, S = Sure). The percentage of the dataset for each level of objectification as well as the average number of concepts per clip are also shown. (Best viewed in colors)} 
  \label{fig_level_factor}
\end{figure}

Second, we analyze the obtained annotations in Fig~\ref{fig_level_factor}. The Sure category is the least represented with 16\%, the Easy Negative being, as expected, the most represented class with 52\% of clips.
It is interesting to note that every concept is approximately annotated with the same rate throughout the Hard Negative, Not Sure and Sure levels of objectification. 

Finally, it is very interesting to observe that the average number of concepts annotated per clip increases with the level of objectification: 1.26 concepts on average per Hard Negative clip, 1.71 for Not Sure, up to 2.6 for Sure. We verify that this trend is observable for every single annotator. It gives an important insight into our video interpretation data: that objectification is a compositional process. This corroborates with recent findings in neuroscience experiments that found that a single element, such as clothing on its own, is not sufficient for people to perceive a character as an object \cite{bernard_revealing_2019}. 

\section{Experiments}\label{sec:exp}

The experiments have two objectives: to verify that the new classification task is feasible, and to identify the challenges of designing efficient models. To tackle these objectives, we consider pre-trained vision models and specifically address the following research questions:
\begin{itemize}
    \item \textbf{Task accuracy --} What are the baseline performances by pre-trained vision models on the objectification detection task? How does the performance vary with hard negative examples?
    \item \textbf{Concept representation --} Can we implement interpretable models of objectification using concepts? What is the quality of representation of every concept, and what are the objectification concepts poorly captured by current models?
\end{itemize}

\subsection{Task accuracy}

\noindent\textbf{Setup}
We discard the Not Sure (NS) class from the \textit{ObyGaze12} dataset, as it gathers by definition samples highly uncertain for humans, and consider the Easy Negative (EN, 62\% of the clip samples), Hard Negative (HN, 19\%) and Sure (S, 19\%) classes.
We approach binary classification in a progressive way, the positive class being made of the S samples. We consider two levels of classification difficulty by composing the negative class either with EN samples, or with HN samples. 
The implementation details of cross-validation and data balancing are provided in Appendix \ref{sec_suppl:task-acc}. 
The average performance over the test set of the best models on validation folds are shown in Table \ref{task-acc} with standard deviations.\\

\noindent\textbf{Baselines} 
We consider video embedding obtained from pre-trained models owing to their zero-shot classification capabilities on video tasks. We select ViViT-B/16 \cite{arnab_vivit_2021}, and the available X-CLIP model, trained on Kinetics \cite{XCLIP}. 

We also re-train a X-CLIP model \cite{X-CLIP} on the LSMDC \cite{rohrbach_movie_2017} film dataset, and refer to Appendix \ref{sec_suppl:LSMDC} for corresponding results, where all implementation details are described.
We keep the pre-trained models frozen and perform an adaptative max pooling of the resulting frame tokens, and feed the output to an MLP made of 2 dense layers, the hidden layer with 128 neurons and ReLU activations, the last with 2 softmax neurons.\\

\noindent\textbf{Results} 
To assess the quality of the models on possibly imbalanced data with a minority of positive samples, we report the F1-scores in Table \ref{task-acc}.
First, by comparing with trivial classifiers (random and an all-positive, see App. \ref{sec_suppl:task-acc}), we observe that \textbf{the task is indeed feasible}, warranted by the data consistency described in Sec. \ref{sec:data} despite the interpretive nature of the task.
Second, we observe that \textbf{the inclusion of Hard Negative examples improves} the classification results, showing the importance of a fine-grained annotation for highly-interpretive tasks.
Results of X-CLIP on other configurations, specifically when the movies of clips in test are different from those in train, are shown in App. \ref{sec_suppl:task-acc}.
The best results based on existing models are of moderate quality, which calls for more investigation into where the difficulties lie.

\begin{table}
\small
\caption{F1-score on the binary task of objectification detection for models trained with easy or with hard negatives and tested on easy or all negative samples, with standard deviations.}
\label{task-acc} 
{\footnotesize
\begin{tabular}{lll|ll}
\toprule
Test           & \multicolumn{2}{c}{EN vs. S} & \multicolumn{2}{c}{(EN U HN) vs. S}\\
Train            & EN vs. S & HN vs. S & EN vs. S & HN vs. S \\
\midrule
ViViT-B/16      & 0.53 (0.18) & 0.62 (0.13) & 0.54 (0.24) & 0.73 (0.1) \\
X-CLIP          & \textbf{0.79} (0.05) & 0.71 (0.05) & 0.66 (0.05) & \textbf{0.82} (0.03)\\
Random          & \multicolumn{2}{c|}{0.32} & \multicolumn{2}{c}{0.28} \\
All positive    & \multicolumn{2}{c|}{0.37} & \multicolumn{2}{c}{0.33} \\
\cmidrule(rl){1-5}

PCBM-DT        & 0.68 & 0.44 & 0.58 & 0.38 \\
PCBM-LR        & 0.64 & 0.43 & 0.50 & 0.37 \\ \hline
\end{tabular}
}
\end{table}

\subsection{Concept accuracy}

To infer on-screen objectification, it is key for the model to detect the means of its production, which correspond to the eight concepts listed in Table \ref{tab-objthesaurus}. We reiterate that in the \textit{ObyGaze12} dataset, every clip annotated with a level of objectification S, NS or HN is also annotated with the presence of instances of the eight concepts. For example, if the \textit{Body} concept is annotated, it means that some level of nudity and/or suggestive body parts are shown on screen, that could contribute to the production of objectification. The means of producing objectification through the eight concepts can be subtle to detect, making it difficult to provide the final interpretation. 
To investigate this difficulty, we implement Post-hoc Concept Bottleneck Models (PCBMs) \cite{yuksekgonul2022posthoc}, which allow us to approach a classification task with pre-trained models in an interpretable way when concept-annotated data is available.
In our case, from the X-CLIP embedding space where our video clips are represented, we identify a Concept Activation Vector (CAV) \cite{Kim2017InterpretabilityBF} for every concept. We then project the X-CLIP embedding of every clip onto the subspace defined by the eight CAVs. The representation of the clip that is the output of this bottleneck is a low-dimensional vector with number-of-concepts components. This vector can then be fed to an interpretable classifier for the objectification detection task.

\paragraph{CAV computation}
For every concept $i$, we collect two sets of samples: positive samples where concept $i$ is present, hence made of S and HN samples with the concept annotated as present, and negative with EN, S and HN without concept $i$. We then train a linear SVM for each concept $i$, the CAV of concept $i$ being the normal vector of the SVM hyperplan.
To train the SVM, we split the data in 10 folds, and reserve the last fold for test. We then perform an 8-fold cross-validation to select the SVM (choice of margin tolerance $c$), every fold training set being balanced with different draws of negative sub-sampling.  

\paragraph{Interpretable classifier}
We then train a decision tree (DT) and a logistic regression (LR) classifier on the same 8-fold cross-validation to classify the level of objectification, the classifiers being fed with the projection of every clip onto the CAVs.\\

\noindent\textbf{Task accuracy with PCBM}
We first verify the quality of objectification detection with F1-scores of PCBM-DT and PCBM-LR shown in Table \ref{task-acc}. The results lower than X-CLIP are expected owing to the known accuracy-interpretablity tradeoff of CBMs \cite{yuksekgonul2022posthoc,zarlenga2022concept}. They are above random and all-positive predictions when training on EN vs. S, which is indicative of the relevance of information held in the concepts.
However, the low results obtained when training the DT and LR to distinguish between S and HN reveals the \textbf{low quality of some CAVs}, where the X-CLIP embeddings cannot be linearly well-separated for these concepts. We investigate this point next.
The resulting DT is discussed in App. \ref{sec_suppl:concept-acc}.\\

\begin{figure}[t]
  \centering
  \includegraphics[width=\linewidth]{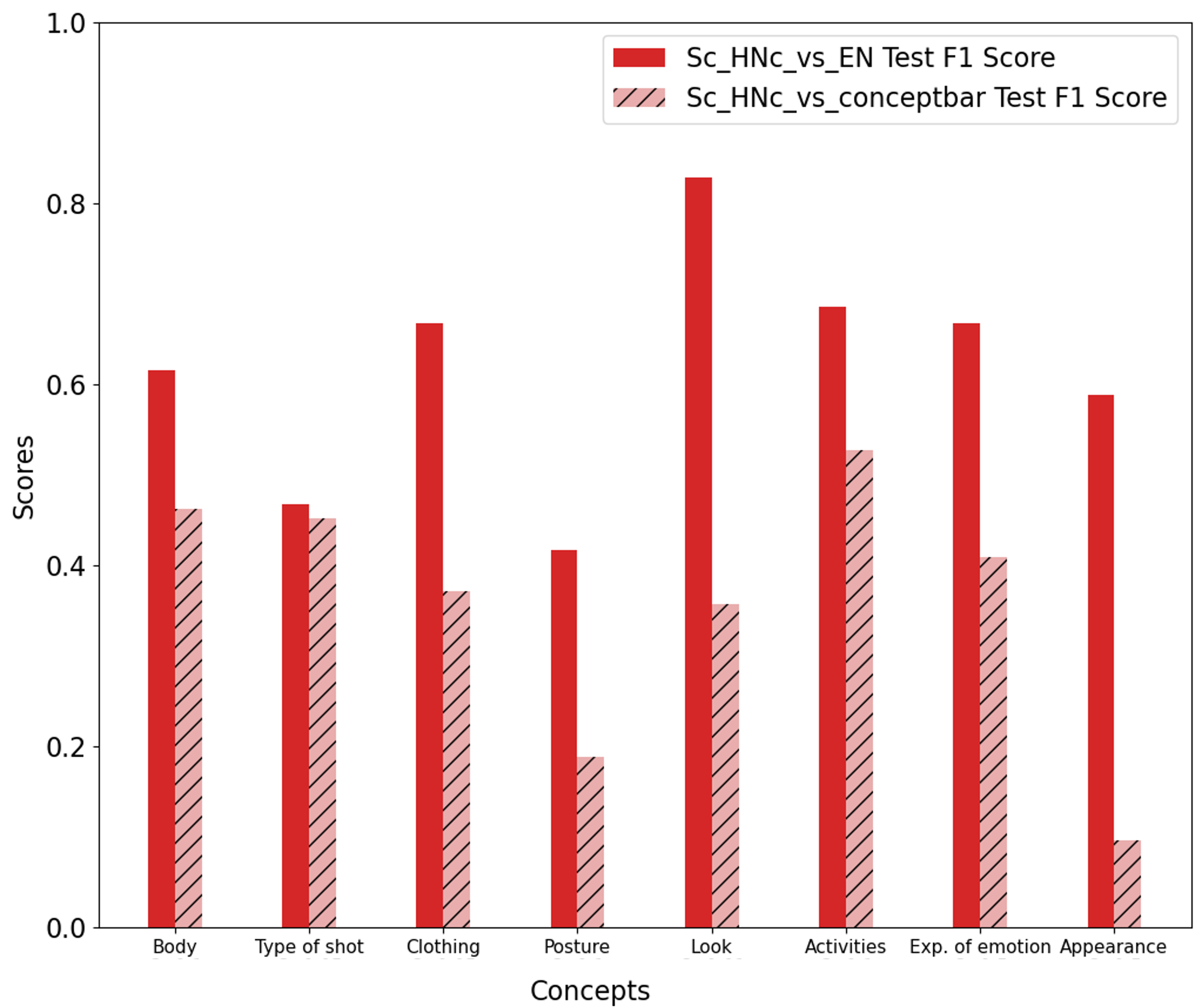}
  \caption{For every concept, F1-score of the best linear SVM selected to define the CAV of this concept. Positive samples (S and HN with the concept) must be separated from: [non-hatched bars] negative samples made of EN only, or [hatched bars] negative samples made of EN and S and HN without the concept. }
  \label{fig_CAVs}
\end{figure}

\noindent\textbf{Concept accuracy}
We now analyze the quality of each obtained CAV by plotting its capability to classify whether the concept is present in a test sample. We consider a positive similarity between the X-CLIP embedding and CAV of concept $i$ indicative of the presence of concept $i$. F1-score on the test set are shown in Fig. \ref{fig_CAVs}. Plots correspond to CAVs obtained from classifying the presence of concept against EN only (solid bars) and against EN with S and HN without concept $i$ (hatched bars). The former is used for PCBM-DT in Table \ref{task-acc}.
We first observe that concept detection is harder when negative samples also include S and HN samples (without the concept). This is expected considering that scenes tagged EN have by definition no element possibly conducive to objectification, and are hence likely to differ visually more from scenes where the concept is present, than do S and HN scenes without the concept.
However within S and HN clips with and without the concept, such shortcuts cannot be exploited anymore. We observe in this case that \textbf{the X-CLIP embedding related to concepts \textit{Type of shot}, \textit{Posture}, \textit{Look} and \textit{Appearance} are harder to separate linearly}. This can be correlated with the analysis of factors of error detailed in App. \ref{sec_suppl:task-acc}. These subtler means of on-screen objectification therefore warrant future work to be properly captured and detected.

\section{Discussion}\label{sec:discu}
\paragraph{Ethical aspect}

This work has an explicit societal motivation in its purpose to tackle, with the help of AI, the analysis of complex temporal patterns operated in cinema that produce the perception of certain characters as objects. This is a challenging but valuable task that aims to uncover and quantify differences in how various identities may be portrayed on screen.

\paragraph{Limitations and challenges}

A distinctive element of our work is the subjective judgement involved in annotating granular video elements for objectification. Video annotation is tedious, and approaching data annotation for such an interpretive task in a rigorous way is even more so, and difficult to scale. We therefore believe that pursuing high-quality, dense annotations with well-defined concepts goes a long way to tackle this new video interpretation task, which represents a valuable new challenge for the computer vision community.

\section{Conclusion}\label{sec:conclu}
In this article, we have introduced a new video interpretation task to detect character objectification in films. We have introduced the \textit{ObyGaze12} dataset, densely annotated by experts for objectification concepts defined from five sub-constructs identified in film studies and psychology. \textit{ObyGaze12} is made available to the community.
We evaluate recent vision models, show the feasibility of the task and where the challenges remain with concept bottleneck models. We show that the representation learning of the concepts of Type of shot, Look, Posture and Appearance need to be improved.

\paragraph{Acknowledgements} This work has been supported by the French National Research Agency through the ANR TRACTIVE project ANR-21-CE38-00012-01 and by EU Horizon 2020 project AI4Media, under contract no. 951911 (https://ai4media.eu/).
{
    \small
    \bibliographystyle{ieeenat_fullname}
    \bibliography{bib_file}
}

\clearpage
\setcounter{page}{1}
\maketitlesupplementary

We make available to the research community the contributed dataset \textit{ObyGaze12}, as well as the code used to produce the results shown in this article and its supplemental material, at \url{https://anonymous.4open.science/r/ObyGaze12/}.

\section{Dataset}\label{sec_suppl:dataset}

This section provides additional information on the list of films, the data annotation and processing procedure, and the calculation of $\gamma$ Inter-Annotator Agreement (IAA).

\subsection{List of films}

The complete list of film of the \textit{ObyGaze12} dataset is shown in Table \ref{tab-films}. It corresponds to a 23\%-subset of the MovieGraphs dataset \cite{vicol_moviegraphs_2018}. The 12 movies we densely annotate for objectification construct and concepts were selected to approximately reproduce the fraction of genres in the original dataset.

\begin{table}
\footnotesize
\caption{Detailed list of films selected for creating dense annotations on objectification, with year of release and genre.}
\label{tab-films}
\begin{tabular}{|l|l|l|}
\hline
\textbf{Film}           & \textbf{Year} & \textbf{Genre(s)}         \\ \hline
Gone Girl               & 2014          & drama, mystery, thriller, \\ \hline
Silver Linings Playbook & 2012          & drama, romantic, comedy   \\ \hline
Crazy Stupid Love       & 2011          & drama, romantic, comedy   \\ \hline
The Help                & 2011          & drama                     \\ \hline
Up in the Air           & 2009          & drama, romantic, comedy   \\ \hline
The Ugly Truth          & 2009          & romantic, comedy          \\ \hline
Marley and Me           & 2008          & drama, family             \\ \hline
Juno                    & 2007          & drama, comedy             \\ \hline
Meet the Parents        & 2000          & romantic, comedy          \\ \hline
As Good As It Gets      & 1997          & drama, romantic, comedy   \\ \hline
Pulp Fiction            & 1994          & drama, mystery          \\ \hline
Sleepless in Seattle    & 1993          & drama, romantic, comedy   \\ \hline
\end{tabular}
\end{table}

\subsection{Data annotation and processing}

\begin{figure}
  \centering
  \includegraphics[width=1.0\linewidth]{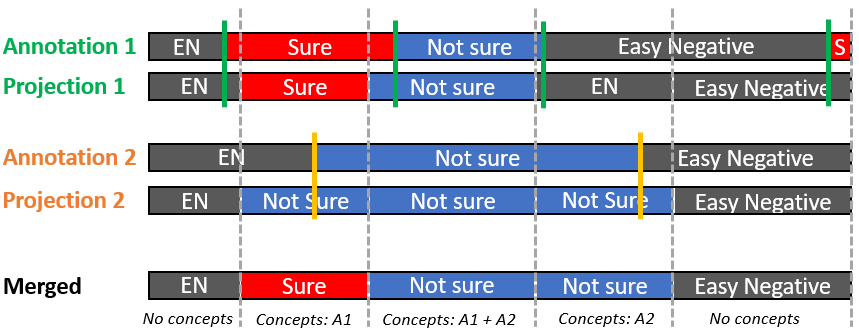}
  \caption{The annotation and data processing procedure is as follows. (1) Two experts annotate each film, with free delimitation (\textbf{\color{Green} Annotation 1} and \textbf{\color{BurntOrange} Annotation 2}). (2) Annotations are projected onto the MovieGraphs delimitation (dashed gray line), taking the highest level of objectification while enforcing a minimum overlap threshold of 20\% (\textbf{\color{Green} Projection 1} and \textbf{\color{BurntOrange} Projection 2}). (3) Projections are \textbf{Merged}, taking the highest level of objectification and merging the concepts only for the same level of objectification.}
  \label{fig:annot-proc}
\end{figure}

The data annotation and processing is illustrated in Fig.~\ref{fig:annot-proc}. During the annotation process, two annotators watch the film, and when they see a scene that is worth annotating, they freely indicate the boundaries of the scene, and then attribute an objectification level as well as concepts, resulting in the \textbf{\color{Green} Annotation 1} and \textbf{\color{BurntOrange} Annotation 2} timelines. Then during the data processing step, the annotations are projected onto the MovieGraphs delimitation (dashed gray lines), taking the highest level of objectification while enforcing a minimum overlap threshold of 20\% (\textbf{\color{Green} Projection 1} and \textbf{\color{BurntOrange} Projection 2}). Annotations that have less than 20\% overlap with the MovieGraphs delimitation are not taken into account  (e.g., clips 1, 3, 4, and 5 of \textbf{\color{Green} Projection 1}), and when multiple annotations have overlap $>$ 20\%, the one with the highest level of objectification is kept (e.g., clips 2 and 4 of \textbf{\color{BurntOrange} Projection 2}). Finally, the projections are \textbf{Merged} to create a single timeline, taking the highest level of objectification and merging the concepts for the same level of objectification.
The reason for this choice is that it appeared in the remediation session that most cases of initial disagreement were scenes that some annotators actually overlooked and agreed the objectification level should be raised to the maximum annotated, also considering concepts they had not noticed at first.

We generated multiple variations of the projections and merge by varying the minimum overlap threshold between 0.1-0.4. As the threshold increases, the numbers of projected and merged clips tagged with Sure and Not Sure decrease while those for Easy and Hard Negative increase, with an overall difference of $[+76,+68,-42,-103]$ clips for the four classes $[EN,HN,NS,S]$. An intermediate threshold of 0.2 was thus chosen for our experiments.

\subsection{Inter-annotator agreement calculation}


The $\gamma$ Inter-Annotator Agreement \cite{mathet2015unified} was designed to address the challenge of annotation tasks on a continuum without pre-defined units. It was motivated by text annotation tasks, but can be equally applied to similar tasks that involve both unitizing and categorization. The calculation reflects this by calculating the score of alignment and category separately:

\begin{equation}
   d^{\alpha,\beta}_{combi(d_{a},d_{c})}(u,v) = \alpha.d_{a}(u,v) + \beta.d_{c}(u,v) \;,
\end{equation}

\noindent where $\alpha$ and $\beta$ represent the weights for the dissimilarities $d_{a}$ and $d_{c}$ in alignment and classification, respectively, for two annotations $u$ and $v$ in the annotation set $\mathcal{A}$. For our work, we calculate the $\gamma$ value on the projected annotations, thus requiring only the $d_{c}$ item (hence $d^{\alpha=0,\beta=1}_{combi(d_{a},d_{c})}(u,v)$) which is defined as a distance matrix between any two objectification levels, that we set to:

\begin{equation}
   d_{c}(u,v) =
\begin{Bmatrix}
 &  EN&  HN&  NS& S  \\
 EN&  0& 0.3&  0.7& 1  \\
 HN&  0.3&  0&0.4  &0.7  \\
 NS&  0.7&  0.4&  0&0.3  \\
 S&  1&  0.7&  0.3& 0
\end{Bmatrix}
\end{equation}

The distances between all annotations in the movie are then averaged to obtain a disorder metric for the entire film:

\begin{equation}
   \delta(a)=\frac{1}{\binom{4}{2}}\cdot\sum_{(u,v)\in \mathcal{A}^{2}}d(u,v)
\end{equation}

In parallel, the dissimilarity is calculated and then averaged over $N$ randomly generated sequences $s$ to obtain a random disorder value for the corpus $\delta(c) = \frac{1}{N}\sum_{s\in c}\delta(s)$. $\gamma$ is then calculated as:

\begin{equation}
   \gamma=1-\frac{\delta(a)}{\delta(c)}\;,
\end{equation}

\noindent where $\gamma\leq 0$ indicates random or worse. For our calculations, we used $N=62$, at which $\gamma$ has a confidence of $p<0.01$.

Over every combination of pairs of annotators per film, we had in total 23 pairs of annotations which achieved an average of $\gamma=0.42$, indicating a moderate level of agreement. Such a level is expected given the interpretive nature of the task and the low number of annotator per data sample. Recent works improve learning approaches by explicitly considering the IAA in cases of low number of annotators with moderate agreement \cite{Wei2021LearningWN, wei_aggregate_2023, bucarelli_leveraging_nodate}. Not considering the clips annotated Not Sure (NS), which is the uncertain and ``noisy'' class in human annotations, the IAA increases to $\gamma=0.69$.

\section{Experiments on task accuracy}\label{sec_suppl:task-acc}

\subsection{Setup details}\label{sec:setup_details}
The implementation details of cross-validation and data balancing used to obtain the results presented in Table \ref{task-acc} are as follows.
For each choice of negative set (EN or HN), each class is split into 10 equal-size folds. The last (resp. last but one) fold of each class is reserved for test (resp. validation), hence preserving class ratios. The remaining 8 folds of the positive class are used for train, while for each remaining 8 folds of negatives, a subset of folds is picked so as to obtain a balanced training set, the number of training sets depending on the class imbalance. The validation set allows to select the best model over training epochs for each training sets. 
The average performance of the models over the test set are shown in Table \ref{task-acc}.

We keep the pre-trained models frozen and perform an adaptative max pooling of the resulting frame tokens, and feed the output to an MLP made of 2 dense layers, the hidden layer with 128 neurons and ReLU activations, the last with 2 softmax neurons.
Experiments were carried out using a GTX 1080 Ti GPU, training of the MLP took approximately 1 hour and inference 30 minutes. Features extraction was performed with a GTX 1080 Ti GPU for 4 hours on average.

\subsection{Random and all-positive baselines}
In Table \ref{task-acc}, we consider two trivial baselines independent of the data sample: \textit{random} predicting positive with probability 0.5, and \textit{random} predicting only positive. In such cases:
\[\mbox{precision} = F_{\mbox{data}}\quad;\quad \mbox{recall} = F_{\mbox{classifier}}\;,\]
where $F_{\mbox{data}}$ is the fraction of positive samples in the test data, and $F_{\mbox{classifier}}$ is the fraction of samples predicted positive by the classifier.
$F_{\mbox{classifier}}$ is 0.5 and 1 for \textit{random} and \textit{all-positive}, respectively.
$F_{\mbox{data}}$ is 23\% and 19\% for test sets EN vs. S and (EN U HN) vs. S, respectively.
The resulting F1-scores are indicated for each trivial baseline and each test set in Table \ref{task-acc}.

\subsection{X-CLIP results on unseen movies versus unseen clips}
In order to assess the feasibility of the task, the results presented in Table \ref{task-acc} are obtained when clips are split randomly between train, validation and test sets, as described in Sec. \ref{sec:setup_details} above. Different clips from the same movie can therefore be in the training and test sets. It is hence possible that the X-CLIP adaptation presented in Table \ref{task-acc} result from overfitting on specific movies. We here test this hypothesis and consider distinct movies between train, test and validation sets. 

We consider 10 movies for possible test and validation sets. Each test set is made of one of these movies. For each test set, validation sets are successively made of one of the 9 remaining movies. For each validation set, the training set is made of the remaining movies in the dataset. 
The training is made considering negative clip examples are HN only. Test is run on (ENUHN) vs. S clips.
Other setup details are kept similar to those used to obtain the results shown in Table \ref{task-acc} and described in Sec. \ref{sec:setup_details}.

For each test set (movie), we present the average (and standard deviation) of the F1-scores obtained over all 9 best models for each validation set. Table \ref{acc_per_testmovie} shows results averaged over all test sets and over every test movie.
We observe that average F1-score is 0.53, to be compared with 0.82 in Table \ref{task-acc}. While the results are still significantly over chance, they also show that the generalization over movies is harder than over clips only, and make for a future challenge to tackle.

\begin{table}[ht]
\small
\caption{F1-score on each movie test, when movies in train, validation and test sets do not overlap.}
\label{acc_per_testmovie} 
\centering
\begin{tabular}{ll}
\hline
\textbf{Test movie} & \textbf{F1-score}\\
\midrule
As Good as it gets & 0.55 (0.09)\\
Crazy, Stupid, Love & 0.59 (0.05)\\
Gone Girl & 0.54 (0.10)\\
Juno & 0.67 (0.09)\\
Marley and Me & 0.55 (0.07)\\
Pulp Fiction & 0.29 (0.13)\\
Silver Linings Playbook & 0.58 (0.03)\\
Sleepless in Seattle & 0.51 (0.06)\\
The Help & 0.56 (0.05)\\
Up in the Air & 0.46 (0.05)\\
\midrule
\textit{Average} & 0.53\\
\bottomrule
\end{tabular}
\end{table}

\section{Experiments on concept accuracy}\label{sec_suppl:concept-acc}

\subsection{Analysis of the decision tree}
\begin{figure*}[t]
  \centering
  \includegraphics[width=0.9\linewidth]{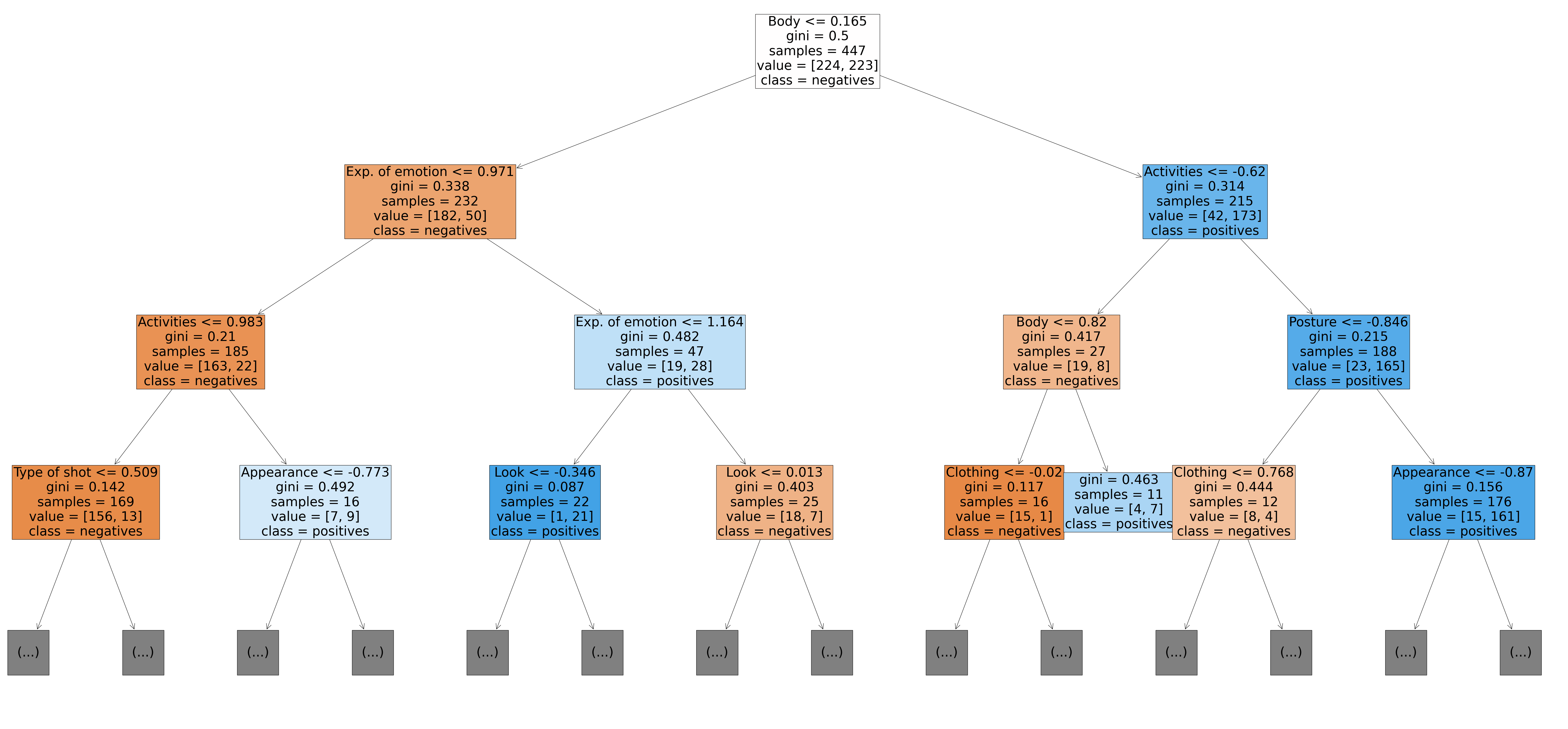}
  \caption{Decision tree trained for the objectification detection task of Easy Negative vs. Sure, fed with embedding similarities to CAV obtained from contrasting clips with concept against Easy Negative examples. Orange (resp. blue) shaded boxes represent a majority of negative (resp. positive) clip examples (i.e., without or with objectification).}
  \label{fig:tree}
\end{figure*}

Here we analyze the decision tree corresponding to the PCBM-DT model shown in Table \ref{task-acc} when the training set in Easy Negative vs. Sure. We remind that this decision tree is fed with the vector of similarities of the X-CLIP embedding of the clip to classify compared with every CAV. The CAV are obtained by training the SVMs on binary classification with negative examples being EN and positive examples being S and HN with the concept, as described in Sec. 4.2.
The decision tree has a depth of 10 and the 4 first levels are shown in Fig. \ref{fig:tree}.

We first observe that a majority of child nodes on the left-hand side of their parent nodes correspond (i) to similarities with concepts lower than a threshold, and (ii) to a majority of negative samples. This is a  consistent result, as the presence of a concept is conducive to a higher probability of an overall rating of objectification. Let us notice that this is not the case for the light-blue node with criterion \textit{Expression of an emotion}, which shows this concept is likely not well captured by the X-CLIP embeddings.
Second, with \textit{Body} as root node, we observe that the presence of concept \textit{Body} tends to structure the construct into two groups of occurrences of objectification: in the left-hand side sub-tree, when the concept tends to be absent, important discriminants are \textit{Expression of an emotion}, \textit{Look}, \textit{Type of shot} and \textit{Activities}; on the right-hand side sub-tree, when the \textit{Body} concept tends to be present, important discriminants are \textit{Posture}, \textit{Clothing}, \textit{Appearance} and \textit{Activities}.

Beyond serving to analyze which concepts are currently poorly captured by existing models, the interpretable classifiers in a PCBM approach also serve film studies experts to analyze whether such groupings can corroborate existing theoretical analyses, or whether it is relevant to expand these analyses thanks to the newly identified groupings.

\subsection{Error analysis}
\begin{figure*}[t]
  \flushleft
  \includegraphics[scale=0.5]{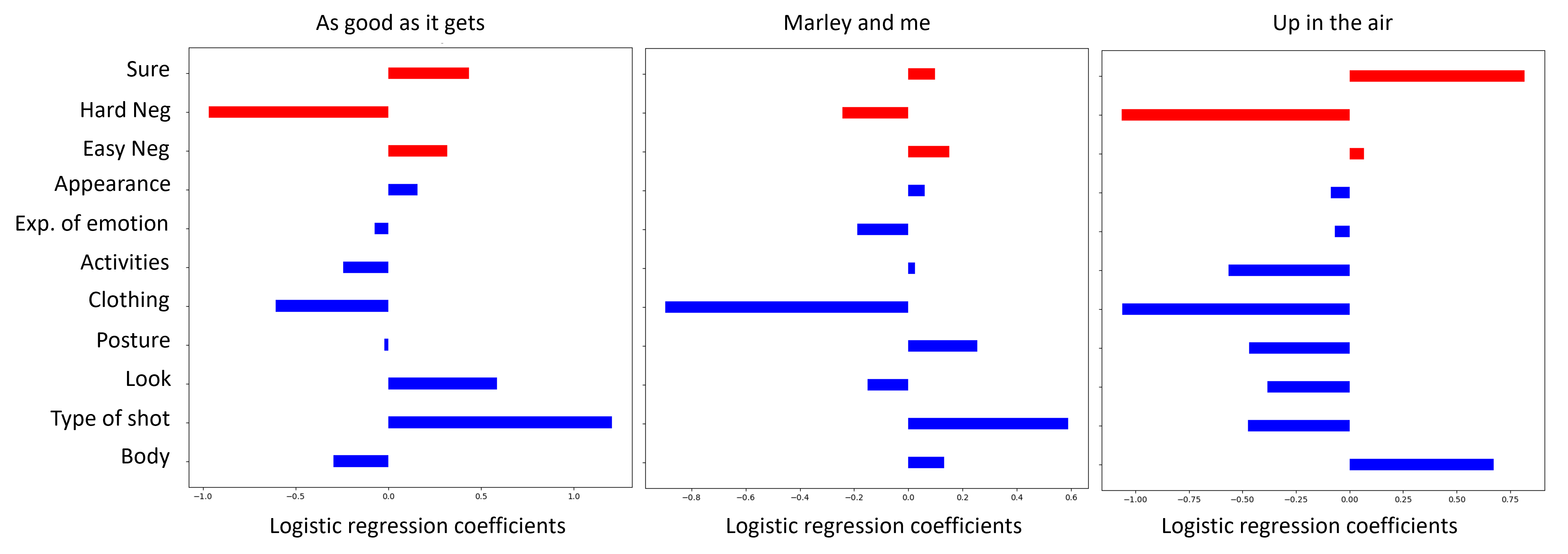}
  \caption{Analysis of the factors of error for the objectification detection task: weights of a logistic regressor predicting whether the test set examples are well classified or not. Positive (resp. negative) weights indicate a positive (resp. negative) contribution to classification success. Left: \textit{As good as it gets}, Middle: \textit{Marley and me}, Right: \textit{Up in the air}.}
  \label{fig:error_factors}
\end{figure*}

In this section we analyze the factors impacting objectification classification errors corresponding to the results shown in Table \ref{task-acc}.
We select an average X-CLIP-based adaptation model trained on HN vs. S and tested on (ENUHN) vs. S, and consider its predictions on the clips in the test set. We label each test clip with 0 if the model fails to predict the correct label of the clip, and with 1 otherwise. We describe the clip with a one-hot encoding vector corresponding to all 11 factors shown the y-axis of Fig. \ref{fig:error_factors}: every of the 8 concepts, and the Sure (S), Hard negative (HN) and Easy negative (EN) labels.
We then train a logistic regression model on these clip descriptors to predict the failure/success labels. 

Results are shown in Fig. \ref{fig:error_factors}, as the regression weights associated with each factor, for three film examples. A negative weight indicates a contribution of the factor to a classification failure.
We first observe that the HN characteristic contributes to a classification failure of the model, while S and EN contribute to classification success. 
Second, we observe that there is variability over the movies on the presence of which concept strongly influences the classification success. However, the presence of the \textit{Clothing} concept seems to be a strong confuser. This can be due to the frequency of appearance of this concept in HN samples, and to the subtlety of the description of this concept (provided in Table \ref{tab-objthesaurus}), which should make it difficult for a pre-trained model to discriminate between an objectifying and non-objectifying overall label on the basis of \textit{Clothing}.
Third, it is worth noting that the concepts shown to be poorly linearly separable when described with the X-CLIP embeddings (see CAV analysis in Sec. 4.2), are also those with a non-stable contribution to the model errors over the film examples: \textit{Type of shot}, \textit{Look}, \textit{Posture} and \textit{Appearance}. 

\section{Experiments with an X-CLIP model pre-trained on LSMDC}\label{sec_suppl:LSMDC}
In complement to the results in Table \ref{task-acc} of the X-CLIP model from \cite{XCLIP} trained on the Kinetics dataset, we train another X-CLIP model introduced by Ma et al. \cite{X-CLIP} on the LSMDC movie dataset, following the procedure described in the code repository of \cite{X-CLIP}. Given the dissimilarity between Internet or instructional videos (such as those of Kinetics) and movies (noted, e.g., by \cite{MovieCLIP}), our objective is to assess whether a model pre-trained on movie videos can achieve better performance at the new objectification-in-movie detection task.
Following the guidelines in the code repository of \cite{X-CLIP}, we retrained the X-CLIP model on the LSMDC dataset from scratch for 5 epochs. We used 4 RTX 8000 GPUs for 5 hours.
Features extraction was performed with a GTX 1080 Ti GPU for 4 hours on average.

Table \ref{task-acc_LSMDC} presents the results of the model, obtained in the same condition as those presented in Table \ref{task-acc}, to be compared with those of X-CLIP \cite{XCLIP} pre-trained on Kinetics. We observe that the results are statistically equivalent, underlying the need for more efficient learning strategies to consider the specific concepts involved in the objectification occurrences. 

\begin{table}[ht]
\small
\caption{F1-score (average with standard deviations) obtained similarly as for Table \ref{task-acc} with the X-CLIP model of \cite{XCLIP} re-trained on the LSMDC movie dataset.}
\label{task-acc_LSMDC} 
{\footnotesize
\begin{tabular}{lll|ll}
\toprule
Test           & \multicolumn{2}{c}{EN vs. S} & \multicolumn{2}{c}{(EN U HN) vs. S}\\
Train            & EN vs. S & HN vs. S & EN vs. S & HN vs. S \\
\midrule
X-CLIP \cite{X-CLIP}   & \multirow{3}{*}{0.70 (0.08)} & \multirow{3}{*}{0.70 (0.10)}  & \multirow{3}{*}{0.66 (0.06)} & \multirow{3}{*}{0.78 (0.11)} \\
pre-trained & & & & \\ 
on LSMDC & & & &  
\end{tabular}
}
\end{table}

\end{document}